\documentclass{IEEEtran}
\usepackage{graphicx}
\usepackage{epsfig}
\usepackage{amsmath}
\usepackage{cite}
\usepackage{booktabs}
\usepackage{multirow}
\usepackage{diagbox}

\usepackage{amsmath,amssymb,amsfonts}
\usepackage{algorithmic}
\usepackage{graphicx}
\usepackage{textcomp}

\usepackage[colorlinks,linkcolor=blue,citecolor=blue]{hyperref}   
\newcommand{\bP}{{\ensuremath{\boldsymbol{P}}}}

\newcommand{\bT}{{\ensuremath{\boldsymbol{T}}}}

\newcommand{\bA}{{\ensuremath{\boldsymbol{A}}}}
\newcommand{\ba}{{\ensuremath{\boldsymbol{a}}}}
\newcommand{\bC}{{\ensuremath{\boldsymbol{C}}}}

\title{\LARGE\textbf{OXSeg: Multidimensional attention UNet-based lip segmentation using semi-supervised lip contours}}

\author{Hanie Moghaddasi\thanks{Corresponding author: hanie.moghaddasi@wrh.ox.ac.uk}{$^1$$^,$$^2$}, Christina Chambers {$^3$}, Sarah N. Mattson {$^4$}, Jeffrey R. Wozniak {$^5$}, Claire D. Coles {$^6$}, Raja Mukherjee {$^7$}, and Michael Suttie{$^1$$^,$$^2$}\\

$^1$Nuffield Department of Women's \& Reproductive Health, University of Oxford, Oxford, United Kingdom \\
$^2$Big Data Institute, University of Oxford, Oxford, United Kingdom\\
$^3$ Department of Pediatrics, University of California San Diego, La Jolla, CA, USA\\
$^4$ Department of Psychology, Center for Behavioral Teratology, San Diego State University, San Diego, California, USA\\
$^5$ University of Minnesota Twin Cities, Minneapolis, Minnesota, USA\\
$^6$ Department of Psychiatry and Behavioral Sciences, Emory University School of Medicine, Atlanta, Georgia, USA\\
$^7$ Faculty of Health and Medical Science, University of Surrey Medical School, Guildford, United Kingdom}
\begin{document}
\maketitle

\begin{abstract}
 Lip segmentation plays a crucial role in various domains, such as lip synchronization, lip-reading, and diagnostics. However, the effectiveness of supervised lip segmentation is constrained by the availability of lip contour in the training phase. A further challenge with lip segmentation is its reliance on image quality, lighting, and skin tone, leading to inaccuracies in the detected boundaries. To address these challenges, we propose a sequential lip segmentation method that integrates attention UNet and multidimensional input. We unravel the micro-patterns in facial images using local binary patterns to build multidimensional inputs. Subsequently, the multidimensional inputs are fed into sequential attention UNets, where the lip contour is reconstructed. We introduce a mask generation method that uses a few anatomical landmarks and estimates the complete lip contour to improve segmentation accuracy. This mask has been utilized in the training phase for lip segmentation.
To evaluate the proposed method, we use facial images to segment the upper lips and subsequently assess lip-related facial anomalies in subjects with fetal alcohol syndrome (FAS). Using the proposed lip segmentation method, we achieved a mean dice score of 84.75\%, and a mean pixel accuracy of $99.77\%$ in upper lip segmentation. To further evaluate the method, we implemented classifiers to identify those with FAS. Using a generative adversarial network (GAN), we reached an accuracy of $98.55\%$ in identifying FAS in one of the study populations.
This method could be used to improve lip segmentation accuracy, especially around Cupid’s bow, and sheds light on distinct lip-related characteristics of FAS. \footnote{This work was supported by NIH grants U01AA014809 (M.S), U01AA014835 (C.C), U01AA014834 (S.N.M), U01AA026102 (J.R.W),
U01AA030164 (J.R.W), and U01AA026108 (C.D.C), as part of the Collaborative Initiative on Fetal Alcohol Spectrum Disorders
consortium.}$^,$ \footnote{This work has been submitted to the IEEE for possible publication. Copyright may be transferred without notice, after which this version may no longer be accessible.}
\end{abstract}
\textbf{Keywords}:
Attention UNet, Fetal alcohol syndrome, Lip segmentation, Mask generation, Multidimensional inputs, Sequential networks

\section{Introduction}
Medical image segmentation has been extensively utilized in the field of computer-aided diagnosis, encompassing applications ranging from magnetic resonance imaging (MRI), computed tomography (CT), and ultrasound to facial images. The facial segmentation primarily focuses on cardinal regions, emphasizing areas such as eyes \cite{rot2018deep}, nose \cite{dibekliouglu2009nasal}, and lips \cite{liew2003segmentation}. More specifically, lip segmentation has a broad application across several domains, including cosmetics, e.g., improving lip wrinkles \cite{ryu2005improving}, speech recognition tasks such as lip synchronization \cite{ma2024decoupled}, and automatic lip-reading (or visual speech recognition) \cite{sheng2024deep} , and the diagnosis of facially affected conditions \cite{suttie2013facial, dixon2011cleft}. The common characteristic of these segmentation applications is their reliance on accurate segmentation boundaries, which can only be attained through different segmentation algorithms tailored to each context.

A wide range of approaches for lip segmentation have been proposed and investigated, each addressing some challenges in facial image segmentation. One commonly used approach involves segmenting lips from the background by leveraging the capabilities of color intensity. While this approach is computationally simple, it relies heavily on image quality, skin tones, color contrast, and brightness and is insensitive to edges and boundaries \cite{eveno2001new}. To enhance the robustness of the color-based segmentation method against the edges, multi-scale wavelet edge detection has been employed to extract lips \cite{guan2008automatic}. Although this method has the advantage of automatic segmentation without relying on a segmentation mask, there remain areas of improvement, particularly in enhancing spatial accuracy and lip vermilion border (demarcation between the lip \cite {lip} and the adjacent skin) detection. Another approach for image segmentation is to utilize model-based techniques \cite{liew2000lip, shdaifat2003active, delmas1999automatic}. However, the accuracy of these methods depends highly on the parameters of the mouth model, and optimal parameters can only be found through a user-guided workflow.

Recent developments in deep learning techniques have made convolutional neural networks (CNN) the backbone of many segmentation algorithms. In image segmentation, fully convolutional networks (FCN) \cite{long2015fully} and UNet \cite {ronneberger2015u} demonstrate the best performance in terms of accuracy and reliability. The UNet consists of two sections. The first section compresses the image into a latent subspace (encoder section). Subsequently, the second section expands the latent to reconstruct the spatial resolution and predict the segmentation mask (decoder section). The interconnection between these sections is gained by skipping connections that reconstruct fine-grained information. While utilizing skip connections helps reconstruct the spatial information, it comes at the cost of creating redundant information in the model, leading to increased computational costs.
To overcome this challenge, Oktay et al. proposed attention UNet to reduce the emphasis on irrelevant areas and emphasize the region of interest \cite{oktay2018attention}.

Several extensions have been introduced to the original UNet to improve its performance, tailored to a specific application. Yang et al. proposed integration of UNet with a fuzzy graph reasoning module to handle image noise and improve boundary detection in lip segmentation. Although it improved segmentation accuracy in images with different backgrounds and noise levels, its application to deal with intense lighting scenarios and reconstruction of the vermilion borderline, especially the cupid’s bow and oral commissures (corners of the mouth), remained limited. Additionally, similar to many of the methods mentioned earlier, its segmentation accuracy depends on the complete segmentation mask in the training phase.

Despite the improvements in the lip segmentation performance, there are two common issues in most of the mentioned methods: 1) Segmentation performance relies on a complete and perfect segmentation mask with a complete contour. This implies the need for a massive amount of complete labeled datasets that are often not readily accessible. 2) Image quality, lighting, contrast, and skin tone contribute to the detection of lip contours. These could serve as potential sources of error that make the boundaries vague, leading to inaccuracies in boundary detection.
This paper presents a method designed to address lip segmentation performance challenges to
\begin{enumerate}
    \item generate a segmentation mask by utilizing only a few initial points while estimating the complete contour by mapping the anatomical landmarks to a lip template.
    \item reduce image quality effects by introducing multidimensional input that can explicitly determine lip boundaries and extract micro-patterns and image texture hidden in the image.
    \item develop a sequential segmentation model that facilitates the segmentation task by refining the boundaries and edges of lips.
\end{enumerate}
To illustrate the performance of the proposed method and demonstrate its application, we applied it to a dataset to assess patients with fetal alcohol syndrome (FAS).
\subsection{Application on fetal alcohol syndrome}\label{FAS}
\begin{figure}[t!]
    \centering
    \includegraphics[width=0.45\textwidth]{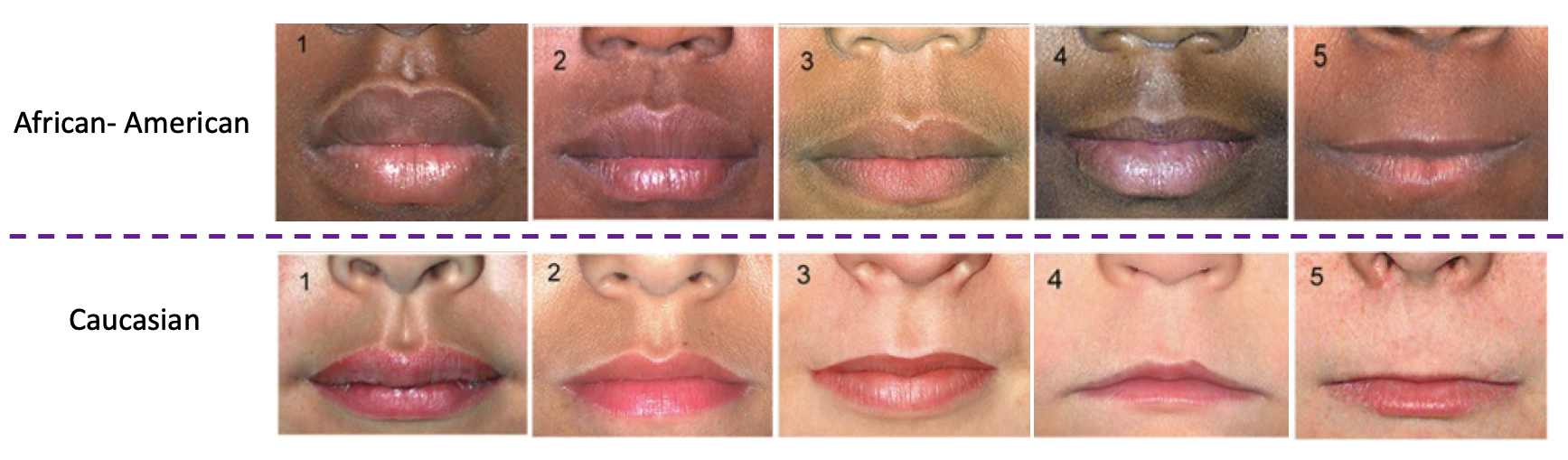}
    \caption{ 5-point Likert scale for lip thickness score ( adopted from \cite{astley2015palpebral}).}
    \label{lip_chart}
\end{figure}

Fetal alcohol spectrum disorders (FASD) is an umbrella term used to describe the spectrum of conditions that arise from the teratogenic effects of prenatal alcohol exposure. Fetal alcohol syndrome (FAS) is one such condition that is clinically identifiable by utilizing four domains: facial anomalies, growth deficiency, deficient brain growth, and neurobehavioral impairment \cite{hoyme2016updated}. Three main facial cardinal features are used to identify those with FAS: 1) short palpebral fissure length (eye width), 2) thin vermilion borders of the upper lip (thin upper lip), and 3) smooth philtrum (indistinct groove of the upper lip) . In this paper, we focus on automated methodology for recognizing the presence of the thin upper lip.
In the clinical environment, a thin upper lip is measured by comparing the lip thickness with a 5-point Likert scale chart to score them between 1 and 5. To address the variation in lip morphology across different ethnic backgrounds, ethnicity-specific charts (shown in Fig. \ref{lip_chart}) have been developed for European and African populations \cite{astley2015palpebral}. On this scale, a vermilion borderline score (VBLS) between 1 and 3 is considered a normal thickness, while a VBLS of 4 and 5 is considered a thin lip.
\begin{figure*}[t]
    \centering
    \includegraphics[width=0.95\textwidth]{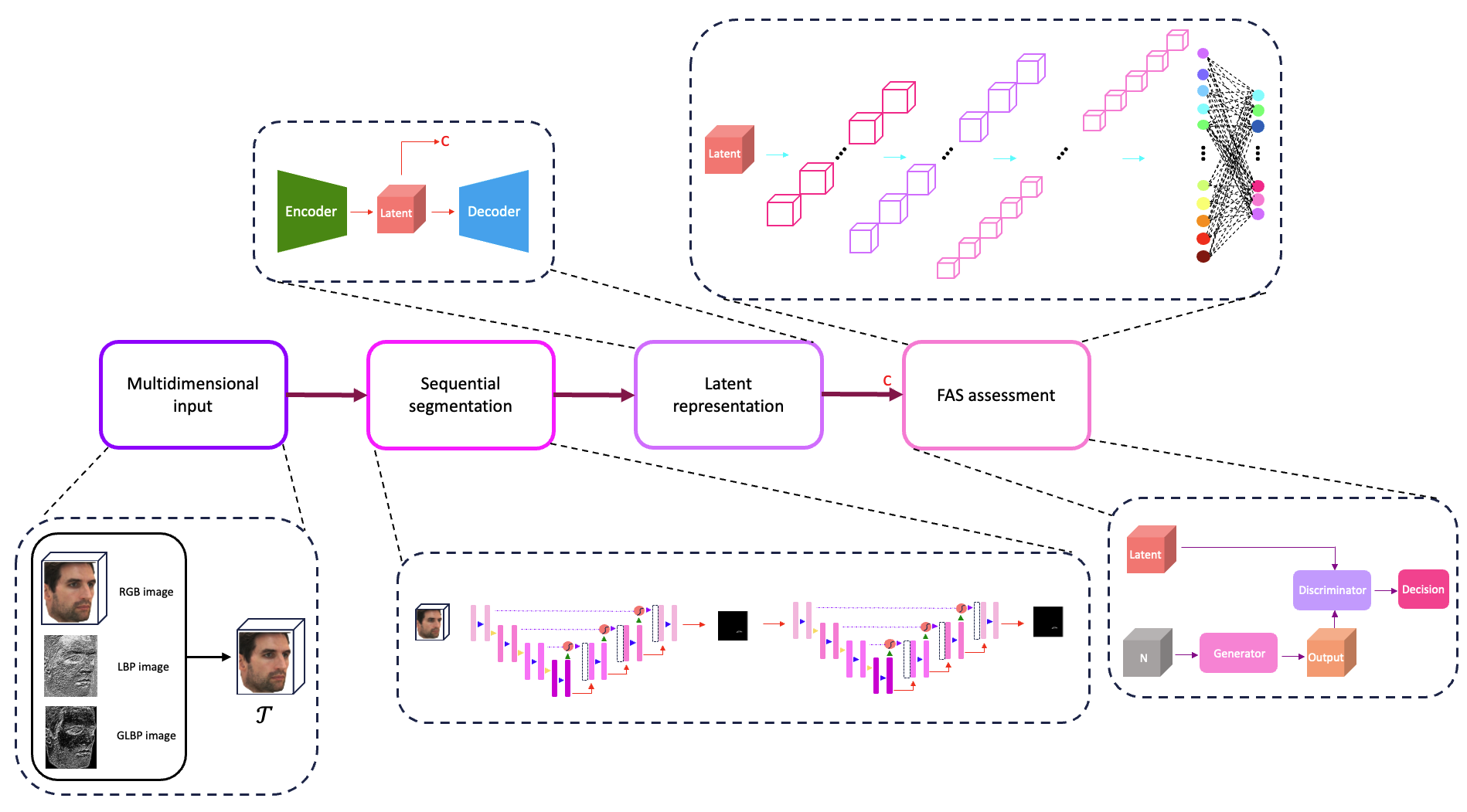}
    \caption{The model architecture. The model takes the RGB images as input, and outputs the FAS status.}
    \label{model_arch}
\end{figure*} 
Clinicians utilize this chart to assess lip thickness. Using the guidelines in Hoyme et al., subjects will meet the facial criteria for a diagnosis of FAS if at least two of the three cardinal characteristics are present. However, the subjective nature of this measurement technique can increase the risk of misdiagnosis and missed diagnoses.  As a result, there is a clinical demand to develop approaches that make the process more objective to improve accuracy and reliability. 

For this purpose, we develop a deep learning-based technique designed first to segment the upper lips and, subsequently, utilize the results to build a model to identify those with FAS. The general block diagram of the FAS identification model is shown in Fig. \ref{model_arch}. We segment the upper lips from raw 2D images in the first two blocks and then utilize the segmented upper lips to construct latent representations. The latent could be used independently by clinicians to assess FAS status (details in Section \ref{discussion}) or transferred to the FAS assessment block where the model distinguishes between FAS or control groups.

The rest of the paper is outlined as follows. Section \ref{method} introduces our method, which includes notation, model architecture overview, multidimensional input, mask generation, sequential segmentation, latent representation, classification, and dataset explanation. Then, lip segmentation and FAS classification evaluation are shown in Section \ref{results}. We discuss latent interpretations and potential future work in Section \ref{discussion}. Finally, the conclusions are drawn in Section \ref{conclusion}.
\section{Methods and algorithms} \label{method}
\subsection{Notation}
This paper uses regular lowercase letters, bold lowercase letters, uppercase letters, bold uppercase letters, and calligraphic uppercase letters for scalars, vectors, 2-tuple, matrices, and tensors, respectively. For example, $a$, $\ba$, $A_i=(x_i,y_i)$, $\bA$ and $\mathcal{A}$ denote a scalar, a vector, a 2-tuple, a matrix and a tensor, respectively. 
$H(.)$ is Heaviside step function and $||.||_F$ denotes the Frobenius norm. 
\subsection{Model architecture overview}
Our approach is structured into four key stages, beginning with processing raw RGB images and leading to assessing the FAS status. The overall high-level block diagram is shown in Fig. \ref{model_arch}. We construct a multidimensional input with the RGB images in the first block. The goal of the second block is to find a contour that minimizes a specific loss function ($\mathcal{L} $). This optimization problem can be written as  
\begin{equation} 
  \theta = \arg\min_{\theta} \mathcal{L}(f_\theta(\mathcal{I}), \bC)
\end{equation}
where $\mathcal{I}  \in \mathbb{R}^{H \times W \times 3}$ is the RGB image with height $H$, width $W$, and three channels. $\bC$ is a binary segmentation mask with the same height and width as the input image, and $f_\theta$ is the model that maps the input image to the segmentation mask. In the next step, we aim to compress the information from the previous step into lower dimensions while keeping the spatial information. We accomplish this by implementing an autoencoder that compresses the segmented upper lips as latent. In the final step, we use compressed data to build classifiers to assess FAS based on the segmented upper lips. 
 
As depicted in Fig. \ref{model_arch}, we focus on the upper lip segmentation phase in the second block. Image segmentation generally starts with using raw RGB (or grayscale) images. However, segmentation accuracy is contingent upon the image quality, brightness, camera characteristics, and spatial resolution. Furthermore, not all regions carry the same level of information uniformly. For example, the most critical regions of interest in lip segmentation are boundaries and edges. To address these limitations, we propose using multidimensional images incorporating regional information.

Section \ref{multi} provides a detailed explanation for constructing multidimensional images, followed by the method to generate segmentation masks in Section \ref{mask}.

\subsection{Multidimensional input} \label{multi}

Multidimensional images are constructed using an image descriptor called local binary pattern (LBP) \cite{ojala1996comparative}. LBP has been predominantly used in facial expression detection \cite{ahonen2006face,ojala2002multiresolution,chao2015facial}, and domains such as cardiac disease diagnostics \cite{moghaddasi2016automatic,moghaddasi2022classification,moghaddasi2024model,yazid2020variable}, and brain MRI analysis \cite{unay2007robustness}, where it is employed to describe image texture. LBP captures the local texture of the image by focusing on the relationship between a given pixel and the neighboring pixels in a predefined mask. This approach unravels micro-patterns in local regions, thereby enhancing image analysis.

\begin{figure}[bt]
    \centering
    \includegraphics[width=0.5\textwidth]{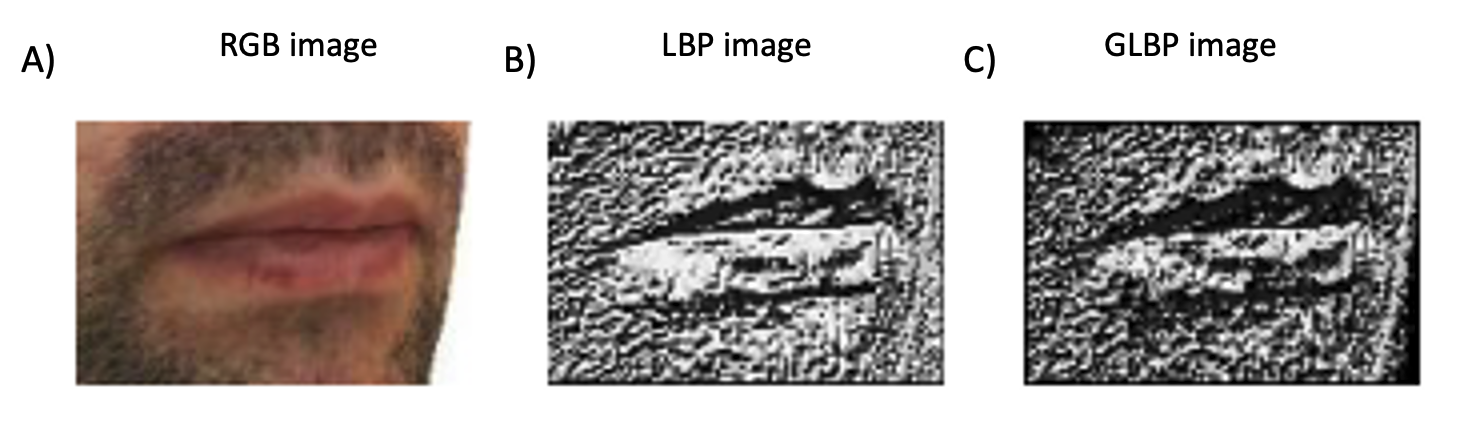}
    \caption{Multidimensional input construction. A) RGB image, B) LBP image, and C) GLBP image. Note that the lip area is zoomed in for visualization purposes.}
    \label{LBP-GLBP}
\end{figure}

The LBP code of a pixel located at $(x_c, y_c)$ is determined by the following calculation \cite{ojala2002multiresolution}:

\begin{equation}\label{lbp1}
  LBP_{P}(x_c,y_c)=\sum_{i=1}^{P} 2^{(i-1)}H\left(I(g_i)-I(g_c)\right)  
\end{equation}

\begin{equation*}
x_i=x_c+Rcos\left(\frac{2\pi i}{P}\right) \\  
\end{equation*}
\begin{equation*}
y_i=y_c+Rsin\left(\frac{2\pi i}{P}\right) \\  
\end{equation*}
where P and R denote the number and radius of neighboring pixels, respectively, $I(g_i)$ is the grayscale value of a pixel at coordinate $(x_i,y_i)$, and $I(g_c)$ denotes the same at the central pixel. In Fig. \ref{LBP-GLBP}. B), the LBP image of a zoomed-in lip area is depicted. Notably, implementing the LBP operator on the image within the lip area has resulted in a more pronounced representation of the boundaries of the upper vermilion border line.

In image segmentation, boundaries and edges are identified as the most significant regions of interest. To emphasize this further, Moghaddasi et al. \cite{moghaddasi2016automatic} introduced an extension of the original LBP that integrates image gradient---which highlights boundaries---and the spatial correlation of the neighboring pixels, resulting in an extensive local binary pattern called gradient LBP (GLBP). 

In GLBP, gradients are computed along horizontal and vertical axes to capture subtle textural variations in the x- and y-directions. Then, the direction of maximum variation is calculated as:

\begin{equation}\label{G_c}
G_c (x_c,y_c)=\frac{g_x(x_c,y_c)g_y(x_c,y_c)}{\max{|g_x g_y|}}
\end{equation}
where $g_x$ and $g_y$ are image gradients along x and y directions, respectively. Then, to further highlight the areas with high gradients, GLBP is computed as follows: 
\begin{equation}\label{Glbp}
  GLBP_{P}(x_c,y_c)=\sum_{i=1}^{P} \left|2^{(i-1)}H\left(I(g_i)-I(g_c)\right)G_c (x_i,y_i)\right|  
\end{equation}

Looking at (\ref{Glbp}), GLBP is calculated using two weights: spatial correlation between pixels (depicted in (\ref{lbp1}) ) and the direction of the high variation in grayscale values of the image (depicted in (\ref{G_c}) ). Together, these two weights help reveal textural micro-patterns in images. In Fig. \ref{LBP-GLBP}. C), the GLBP image is shown. As can be seen, by implementing GLBP, oral commissure and the boundaries of the lower part of the upper lip vermilion became increasingly evident.

Therefore, to incorporate image micro-patterns and subtle textural changes, we construct a tensor $\mathcal{T}$ as follows:
\begin{equation}
\mathcal{T}=\left[\mathcal{I} | LBP | GLBP\right]
\end{equation}
where $\mathcal{I} \in \mathbb{R}^{H \times W \times 3} $, $LBP \in \mathbb{R}^{H \times W } $, $GLBP \in \mathbb{R}^{H \times W } $, and $\mathcal{T} \in \mathbb{R}^{H \times W \times 5}$. We use $\mathcal{T}$ as input for the next phase.
\subsection{Segmentation mask} \label{mask}
Supervised segmentation tasks require a ground truth segmentation mask. The initial anatomical landmarks are obtained in our dataset using the method proposed in \cite {fu2021facial}. The automatically extracted anatomical landmarks are subsequently manually corrected by one of the authors (M.S) to improve the accuracy of the ground truth.

One common approach involves generating heat maps through 2D Gaussian kernels for each landmark. However, this approach results in landmark detection, while in lip thickness analysis, we need to have a complete upper lip contour to analyze the lip area and thickness. To address this problem, we propose to estimate the lip contour from the anatomical landmarks using a lip template.

To find a contour with anatomical landmarks that can serve as the segmentation mask, we start by determining corresponding points on the template by solving the minimization problem as follows:

\begin{equation}
  \arg\min_{\bT} \left|\left| \bT - \bP \right|\right|_F^2
\end{equation}
where $\bP=\{ P_1, P_2,\cdots, P_N\}$ denotes the coordinates of the anatomical landmarks, $N$ denotes the total number of the anatomical landmarks, and $\bT =\{ T_1, T_2,\cdots, T_N\}$ shows the corresponding points of the anatomical landmarks on the lip template. In the next step, we discretize the template to generate more points between the corresponding points. To do so, we parameterize the segment between two consecutive points (i.e., $T_i$ and $T_{i+1}$) to find new points as follows:

\begin{equation}
    T'= \arg\min_{(x_a,y_a) \in \mathcal{C}} \left|\left| (1-a)T_i+aT_{i+1} - (x_a,y_a)\ \right|\right|^2 
\end{equation}
where $\bT' =\{ T'_1, T'_2,\cdots, T'_K\}$ are interpolated points on the template contour $\mathcal{C}=\{\bT,\bT' \}$, $T_a'=(x_a,y_a)$ is the 2D coordinate of a new point on the contour, and $a \in [0,1]$ is the interpolation parameter.

In the next step, we plan to project the interpolated points on the template to the landmark trajectory where they satisfy two conditions:
\begin{enumerate}
    \item These points have a minimum distance to the anatomical landmarks.
    \item The difference between the proportional distance of the new points on the anatomical landmark trajectory to two consecutive anatomical landmarks and the proportional distance of the interpolated points on the template to two consecutive corresponding points should be minimal.
\end{enumerate}

These two conditions ensure the preservation of the object's shape in the new interpolated anatomical landmarks. Therefore, to find these interpolated anatomical landmarks, we define a minimization problem as

\begin{equation}
\begin{array}{rrclcl}
  \arg\min_{A_j} ||P_i-A_j ||^2 \\
  \text{s.t.} & \arg\min_{A_j} \left|\frac{\left|T_i-T'_j\right|}{\left|T_{i+1}-T'_j\right|}-\frac{\left|P_i-A_j\right|}{\left|P_{i+1}-A_j\right|}\right|\\
  & \\
  & \forall i \in 1,2, \cdots, N-1 \\
  & \forall T'_j \in T_i<T'_j<T_{i+1}  
\end{array}
\end{equation}
where $\bA =\{ A_1, A_2,\cdots, A_J\}$ are the interpolated anatomical landmarks. Given anatomical and interpolated landmarks, we connect consecutive points to make a contour that will be used as the segmentation mask.

\subsection{Sequential segmentation}

In Section \ref{multi}, we generated multidimensional inputs, and in Section \ref{mask}, we produced lip segmentation masks, both of which are used to train a segmentation model in this step.

In medical image segmentation, fully convolutional networks (FCN) \cite{long2015fully} and UNet \cite{ronneberger2015u} have been widely employed across various studies like myocardial segmentation\cite{kim2020automatic}, lung cancer analysis \cite{shaziya2018automatic}, and fetal brain analysis \cite{salehi2017auto}. However, these methods are unbiased across different regions of an image, resulting in redundant low-level features and limitations in extracting regional distinctions. To address this problem in image segmentation, Oktay et al. \cite{oktay2018attention} proposed an extension of the original UNet, namely, attention UNet (AUNet). This method highlights regions of interest by using additive soft attention, avoiding irrelevant regions, and decreasing redundant information. 

In this paper, we propose an integration of LBP, and GLBP images, and AUNet. This approach improves the effectiveness of the LBP and GLBP highlighted regions in conjunction with the attention process, resulting in improved segmentation accuracy. However, multidimensional input results in excessively detailed information, leading to redundant low-rank features across the dimensions. To address this problem, we propose to use sequential AUNet on a pre-segmented image. This approach extracts micro-patterns using the first AUNet, and the second one is more responsible for refining the edges and boundaries on the vermilion borderline. Therefore, as a result of the preceding steps, we segmented the vermilion borderline with a sequential segmentation approach. 

The network takes the $\mathcal{T} \in \mathbb{R}^{256 \times 256 \times 5}$ tensor as input. The encoder section consists of four convolutional layers with 64, 128, 256, and 512 filters, each followed by a max pooling layer with a size of $2\times 2$. A bottleneck follows the encoder section. In the decoder section, we used four transposed convolutional layers with 512, 256, 128, and 64 filters for upsampling. Each layer is followed by an attention layer, which generates an attention map to weigh different regions, highlighting the most informative components. Finally, a convolutional layer with a sigmoid activation function outputs a segmented $256\times256$ image. The segmented mask is fed to the second AUNet with the same architecture, resulting in a segmented image of the same size.

\subsection{Latent representation}
In the segmentation phase, we used full-face images, resulting in irrelevant regions on the segmented images and increasing computational costs. To address this issue, we compress segmented images to exclude irrelevant information while preserving important regions of interest. In lip segmentation, the region of interest is the lip area. Therefore, the objective is to compress the image to preserve this area while reducing irrelevant regions. Here, we use an autoencoder \cite{hinton2006reducing} that compacts the segmented images and reduces the dimension. In the autoencoder, the encoder part is responsible for generating low-dimensional latent. Our motivation for utilizing autoencoder for this purpose is twofold: \textit{i)} Compressing images into lower-dimensional latent significantly facilitates the classification task, making it substantially more efficient.  \textit{ii)} The latent representation could be used as a clinical tool for clinicians to assess FAS status. Following this approach, we can find a stereotype latent representation for the FAS group and determine the affected regions of the face (more details are explained in Section \ref{discussion}).

The network takes 2D segmented masks with a dimension of $256 \times 256$. In the encoder section, we employed two convolutional layers with 32 and 64 filters, each followed by a 2D max pooling layer with a size of $2\times2$. In the decoder section, we used two convolutional layers with 64 and 32 filters, each followed by a 2D upsampling layer with a size of $2 \times 2$. Therefore, the latent dimension is $64\times64\times64$.

\subsection{Classification}
To investigate the application of the proposed lip segmentation method, we utilized the extracted latent to assess fetal alcohol syndrome. The latent can be utilized in different approaches for discrimination. One common approach is to flatten the latent (i.e., reshaping the original 3D latent into a one-dimensional vector) that can subsequently be fed to a classifier (e.g., support vector machines or a linear discriminant analysis) that discriminates between FAS and control. While this approach is computationally fast, it does not preserve the spatial information in the latent. To account for spatial information, we employed two methods, a 3D convolutional neural network (CNN) \cite{ji20123d} and generative adversarial nets (GAN) \cite{goodfellow2014generative}, which utilize 3D latent as input and classify between control and FAS. The implementation details are provided in Sections \ref{CNN} and \ref{GAN}.

\subsubsection{3D CNN}\label{CNN}
The network takes a latent with a dimension of $64 \times 64 \times 64$ as input. We employed three 3D convolutional layers followed by 3D max pooling layers. The number of filters for each of the 3D convolutional layers is 32, 64, and 128, respectively, each with a kernel size of $3 \times 3 \times 3$. The ReLu function has been used as an activation function. The kernel size in the 3D max pooling layers is $2 \times 2 \times 2$. Subsequently, the output is flattened and connected to a fully connected layer with 512 neurons, followed by a dropout layer with a ratio of 0.5, to reduce overfitting.
\subsubsection{GAN}\label{GAN}
GAN consists of two models: generative and discriminative models. The generative model starts with random noise to mimic real data, and the discriminator model decides whether the generated data are real or fake. The generator has three convolutional layers with 256, 128, and 64 filters. The discriminator consists of four convolutional layers with 32, 64, 128, and 256 filters. We used a sigmoid neuron for binary classification (FAS or control). 

\subsection{Dataset}
\begin{figure}[bt]
    \centering
    \includegraphics[width=0.5\textwidth]{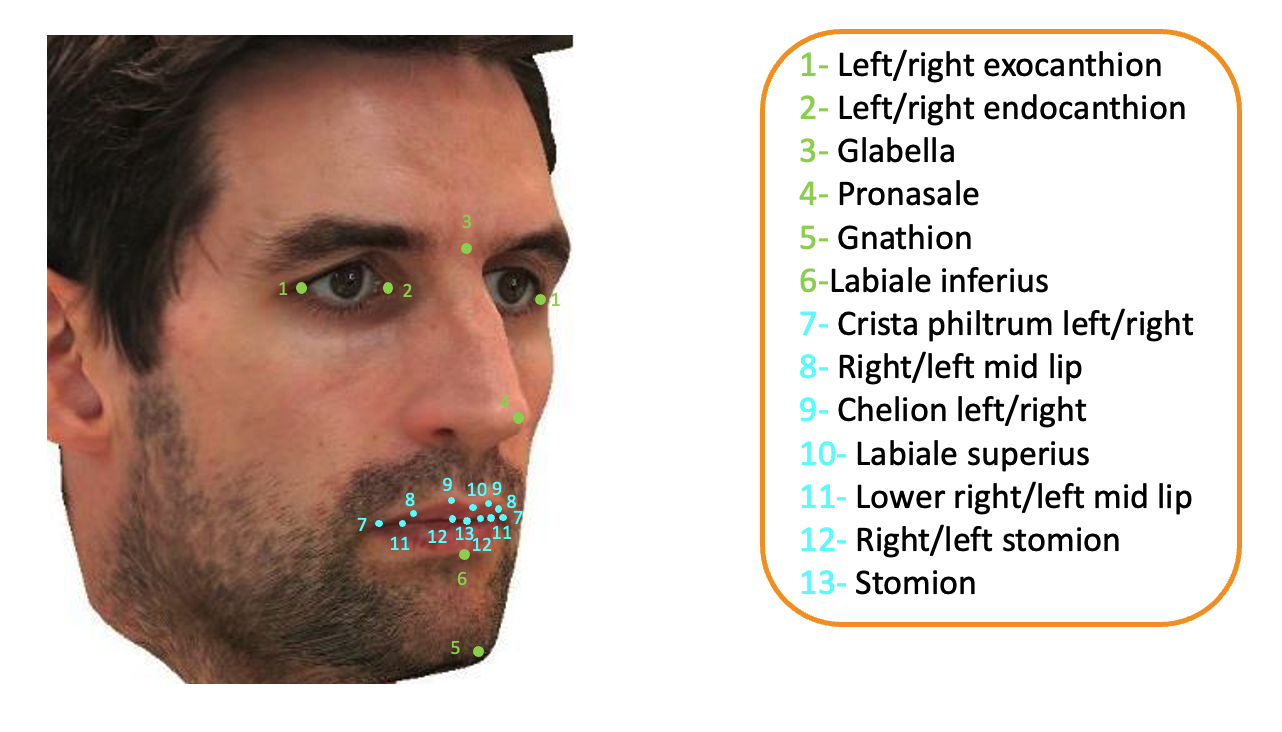}
    \caption{Overlay of the anatomical landmarks on the facial regions. Blue depicts the upper lip landmarks, and green denotes other landmarks used for image alignment.}
    \label{landmarks}
\end{figure}

The Collaborative Initiative on Fetal Alcohol Spectrum Disorders (CIFASD) is a multidisciplinary consortium that focuses on improving the prevention, diagnosis, and treatment of FASD. We utilize high-resolution 3D facial images of $1023$ subjects, with ages between 2 and 20 years, collected from multiple CIFASD sites across the USA. Images were acquired using static-tripod-mounted stereophotogrammetry camera systems (3DMD), which capture $180^\circ$ images of the face, with a geometric resolution of $<0.2mm$. 
We obtained 2D images from portrait-rendered screenshots of the original 3D images for the FAS assessment. Anatomical landmarks, shown in Fig. \ref{landmarks}, are initially extracted by the method explained in \cite{fu2021facial} and corrected by one of the authors to increase accuracy.

We used the complete dataset for lip segmentation, but a subset had missing FAS status and vermilion borderline scores. Consequently, we excluded these subjects for the classification phase. The resulting dataset (n=$453$) had known diagnostic categorizations labeled as either FAS (n=$82$) or control (n=$371$). These subjects had undergone FASD assessments by expert dysmorphologists and neurobehavioral specialists from CIFASD. Subjects were labeled as FAS if they met the criteria for FAS or partial FAS (pFAS) according to the Hoyme criteria\cite{hoyme2016updated}, requiring at least two of the three cardinal facial features. Control subjects were those who did not meet the criteria for any FASD diagnosis and had no reported prenatal alcohol exposure. We excluded any subject with a known genetic condition. $880$ subjects had known VBLS scores, graded between one and five.

\begin{figure}[bt]
    \centering
    \includegraphics[width=0.5\textwidth]{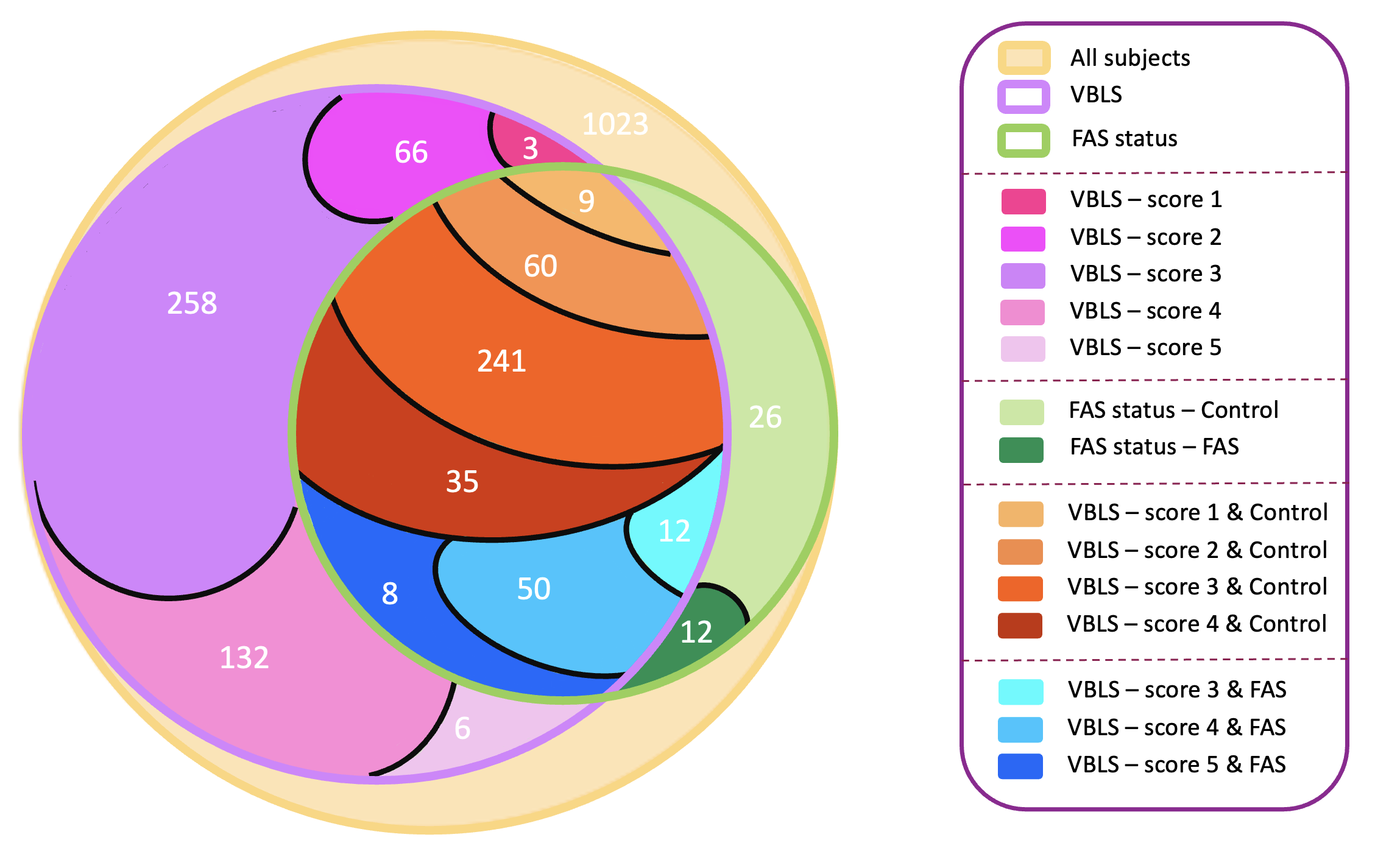}
    \caption{Venn diagram for the data distribution. The pink spectrum denotes the VBLS group, and the green spectrum shows subjects with a known FAS status. The VBLS and FAS status intersection is then subdivided into two groups: control and FAS. The VBLS intersection control is denoted with an orange spectrum, and the VBLS intersection FAS is depicted with a blue spectrum. }
    \label{Venn}
\end{figure}

A detailed Venn diagram is depicted in Fig. \ref{Venn}, to illustrate the data distribution in different subgroups. VBLS denotes subjects with an available vermilion borderline score, and FAS status refers to subjects with an available FAS diagnostic status. Subjects overlapping VBLS and FAS status are subdivided into two groups: control and FAS. In each group (i.e., either control or FAS), the subjects are graded between $1$ and $5$ based on their VBLS. Note that in the control subgroup of the FAS status group, there is no subject with a VBLS of $5$. Similarly, in the FAS subgroup of the FAS status group, there is no subject with 
a VBLS of $1$ or $2$. 

In the classification phase, we performed data augmentation prior to multidimensional input construction to improve the model generalization and increase diversity in our dataset. Given the segmentation goal (i.e., the lip positioned horizontally in the image), we employed horizontal flip, rotation ($5^\circ$), and brightness change (0.8 and 1.1).

In the clinical environment, the vermilion borderline is scored using two different scales for Europeans and Africans (shown in Fig. \ref{lip_chart}). To enhance the robustness and precision of the classification model, we also developed distinct classification models for these ethnicities. 
\section{Results} \label{results}
The results are structured into two main areas: lip segmentation evaluation and FAS assessment evaluation, each highlighting distinct aspects of the study. Section \ref{lip_seg} covers lip segmentation performance, including an illustration of the proposed methods' segmentation results and a quantitative comparison between them using evaluation metrics. Section \ref{classification} focuses on the classification performance using two classifiers, namely, 3D CNN and GAN.

\begin{figure*}[t]
    \centering
    \includegraphics[width=1\textwidth]{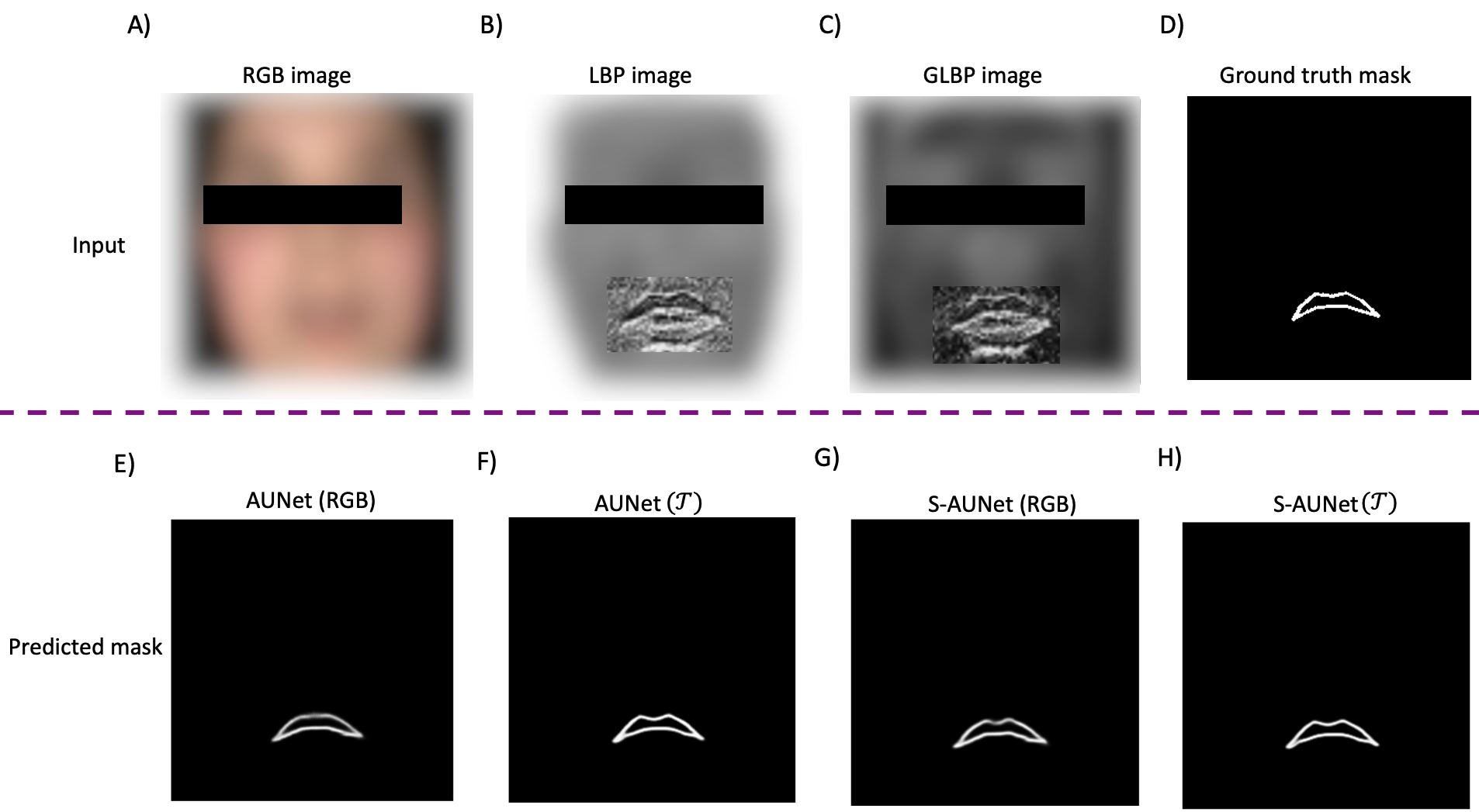}
    \caption{Comparison of predicted mask results across four segmentation methods on a sample image from the test dataset from the European population.}
    \label{4segmentation_w}
\end{figure*}

\begin{figure*}[t]
    \centering
    \includegraphics[width=1\textwidth]{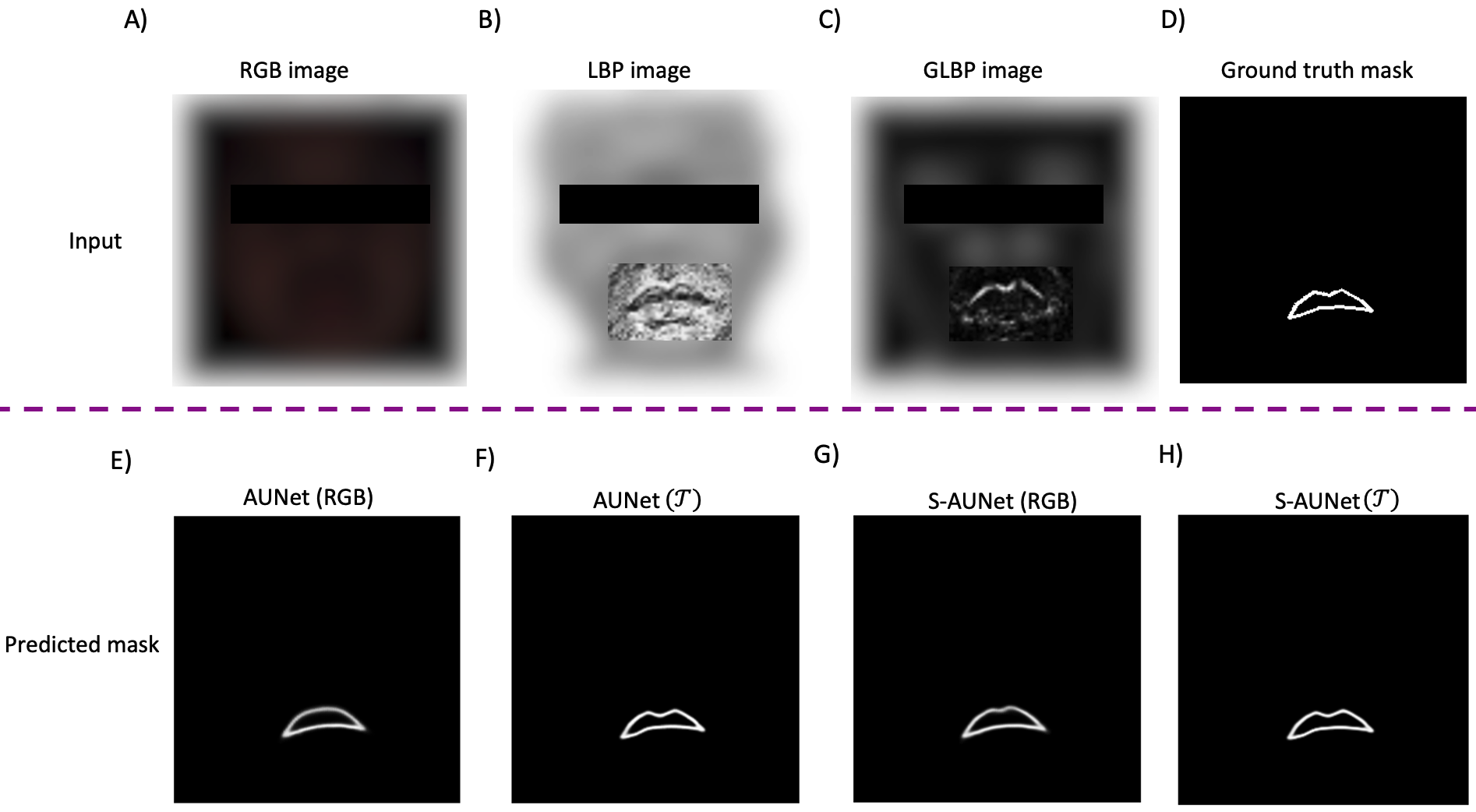}
    \caption{Comparison of predicted mask results across four segmentation methods on a sample image from the test dataset from the African population.}
    \label{4segmentation_AA}
\end{figure*}

\subsection{Lip segmentation evaluation}\label{lip_seg}
We visually investigate the effectiveness of the proposed segmentation method in Fig. \ref{4segmentation_w} and Fig. \ref{4segmentation_AA} for European and African populations, respectively. In the first row of Fig. \ref{4segmentation_w}, the RGB, LBP, and GLBP images are shown in Fig. \ref{4segmentation_w}. A), Fig. \ref{4segmentation_w}. B), and Fig. \ref{4segmentation_w}. C), respectively. Looking at the lip region, the upper boundaries of the vermilion borderline became more distinct using the LBP operator. In contrast, the oral commissures and the boundaries of the lower part of the upper lip became more distinguishable using the GLBP operator. In Fig. \ref{4segmentation_w}. D), the ground truth of the upper lip mask, constructed by the method explained in \ref{mask}, is shown. The second row illustrates the predicted masks estimated by different methods. In Fig. \ref{4segmentation_w}. E), it is evident that when utilizing AUNet with raw RGB images, cupid's bow (curve of the upper lip), chelion (corners of the mouth), and labile superius (upper-lip midpoint) cannot be reconstructed on the predicted mask, highlighting the limitation of raw RGB images. Sequential AUNet (S-AUNet) partially addresses this problem, as shown in Figure \ref{4segmentation_w}.G. However, the cupid's bow still needs to be completely reconstructed. Integrating multidimensional inputs (illustrated in Fig. \ref{4segmentation_w}. F)) and a sequential AUNet enables us to predict the complete vermilion borderline, as depicted in Fig. \ref{4segmentation_w}. H). We observe the same pattern for a subject from the African population in Fig. \ref{4segmentation_AA}. Comparing Fig. \ref{4segmentation_AA}. G) and Fig. \ref{4segmentation_AA}. H) with Fig. \ref{4segmentation_w}. G) and \ref{4segmentation_w}. H), the improvement in reconstructing cupid's bow, chelion, labile superius, and stomion (midpoint between upper and lower lips) using S-AUNet ($\mathcal{T}$) is more pronounced for the African population.

In addition to the visual interpretations, we used six metrics to evaluate and compare the segmentation models quantitatively. The first metric is the Dice score, calculated as
\begin{equation}
    D (X,\hat{X})=\frac{2|X \cap \hat{X}|}{|X|+|\hat{X}|}
\end{equation}
where $X$ is the ground truth segmentation mask and $\hat{X}$ is the predicted segmentation mask. There is a related yet stricter metric known as intersection over union (IoU), which can evaluate the segmentation models in terms of their sensitivity to slight dissimilarities in the intersection between the ground truth and the predicted mask. The IoU can be calculated as 

\begin{equation}
    IoU(X,\hat{X}) = \frac{|X \cap \hat{X}|}{|X \cup \hat{X}|}
\end{equation}
In case of a perfect match between the predicted mask and the ground truth, both the Dice score and the IoU achieve a value of 1. A related metric to the IoU is volumetric overlap error (VOE), defined as 
\begin{equation}
    VOE(X,\hat{X}) = 1- \frac{|X \cap \hat{X}|}{|X \cup \hat{X}|}
\end{equation}

In shape analysis, there is an evaluation metric to measure the maximum discrepancy between two shapes, called Hausdorff distance (HD). This metric can be used in image segmentation tasks to measure the similarity between the ground truth and predicted masks. To find the maximum discrepancy between two sets, HD is defined as 
\begin{equation}
    HD (X,\hat{X})=\max \left\{d_h(X,\hat{X}),d_h(\hat{X},X)\right \}
\end{equation}
\begin{equation*}
    d_h(X,\hat{X})=\max_{x \in X}\min_{\hat{x} \in \hat{X} } ||x-\hat{x}||
\end{equation*}
We calculate the closest point on the predicted mask for each point on the ground truth mask, and the maximum distance within the closest pairs is computed by $d_h(X,\hat{X})$. Finally, considering all points, the greatest distance between two sets is determined as HD. Ideally, lower HD values indicate a lower spatial discrepancy, considering both shape and size.

We also calculate pixel accuracy as
\begin{equation}
    PA=\frac{TP+TN}{TP+TN+FN+FP}
\end{equation}
where TP and TN denote true positives and negatives, respectively, and FP and FN show false positives and negatives, respectively. The pixel accuracy is a metric that considers all classes with a similar weight. However, a semantic segmentation with a class imbalance (i.e., upper lip as the region of interest and the rest of the face as background) needs a more precise metric for evaluation. Therefore, we calculate a more specific metric to measure the accuracy for lip segmentation solely by considering two classes: upper lip and background. The metric is referred to as pixel accuracy per class (here, that is, the upper lip) and computed as
\begin{equation}\label{pac}
 PA_c=\frac{TP_c}{TP_c+FN_c}   
\end{equation}
where $TP_c$ denotes pixels correctly classified in the upper lip class and $FN_c$ shows those classified in the background class but belong to the lip class. In (\ref{pac}), the denominator denotes the total pixels in the upper lip class.

\begin{figure*}[t]
    \centering
    \includegraphics[width=1\textwidth]{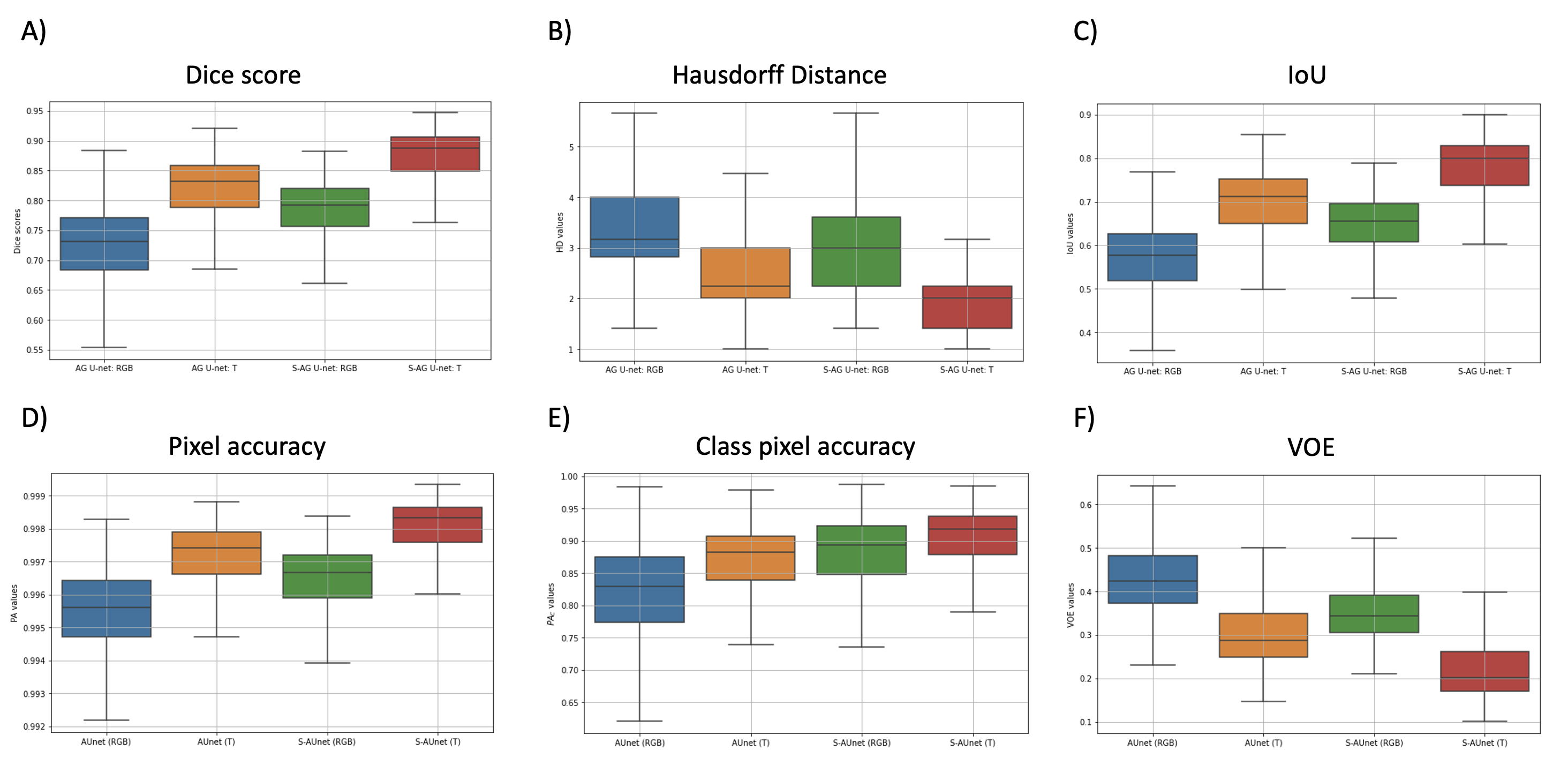}
    \caption{Dice score, Hausdorff distance, IoU, pixel accuracy (PA), pixel accuracy per class (PA$_c$), and VOE box plots. Outliers are not shown on the plots.}
    \label{Dice}
\end{figure*}

These evaluation metrics are calculated for all four segmentation methods as illustrated in Fig. \ref{Dice} and Table. \ref{stat}. The box plots show the distribution of the metrics, for all subjects. As can be seen, the integration of multidimensional input and S-AUNet outperforms the other segmentation models in all metrics. More specifically, the S-AUNet($\mathcal{T}$) model approached closer values to 100\% (representing score 1) in Dice score, IoU, pixel accuracy, and class pixel accuracy, indicating a more accurate match to the ground truth mask. Furthermore, the model also reached lower HD values, demonstrating that the maximum deviation between the contours of the predicted masks and the ground truth masks in S-AUNet ($\mathcal{T}$) is lower than other models. In addition, using the S-AUNet ($\mathcal{T}$) model, we reached lower VOEs compared to other models, indicating a lower error in volumetric overlap between predicted and ground truth masks.

\subsection{FAS classification evaluation} \label{classification}

\begin{figure}[t]
    \centering
    \includegraphics[width=0.5\textwidth]{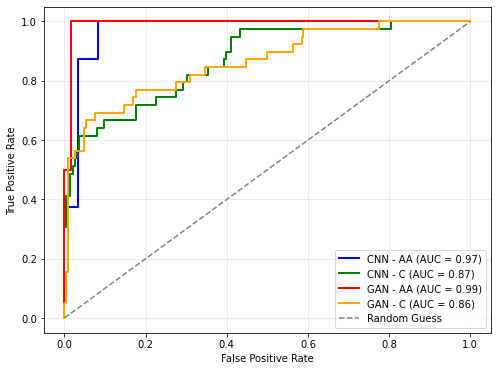}
    \caption{ROC curve}
    \label{ROC}
\end{figure}
To evaluate the segmentation model's performance and demonstrate one application of the segmented lip, we implemented two classifiers, 3D CNN and GAN, to assess discrimination accuracy between FAS and controls. For the FAS assessment, we employed the proposed S-AUNet ($\mathcal{T}$) method for the upper lip segmentation. Since the thickness of the upper lip is scored based on two different scaling systems for Africans and Europeans, we also developed two separate models for these populations. To avoid overfitting, we split the dataset into two subsets, training and testing datasets, with a ratio of $0.2$ for testing.

The classification results, including the train and test loss, along with accuracy, are drawn in Table \ref{loss_acc}. Both classifiers reached an accuracy of over $90\%$ in classifying the test dataset. Using both classifiers, the test accuracy in the African group is higher than in the European group. More specifically, the GAN classifier achieved an accuracy of $98.55\%$ when distinguishing between FAS and the control group. Comparing the classifiers, GAN slightly outperforms 3D CNN. In particular, in African populations, the GAN classifier reached an accuracy of $98.55\%$, while the 3D CNN reached an accuracy of $95.65\%$. Similarly, in European populations, the GAN classifier achieved an accuracy of $92.45\%$, while the 3D CNN achieved an accuracy of $90.56\%$. 

Since we have an imbalanced dataset, we have reported different metrics for a comprehensive evaluation, considering the majority and minority groups. More specifically, we calculated accuracy, sensitivity, specificity, precision, and F1- score as explained in \cite{moghaddasi2022classification}. The classification performance of the 3D CNN and GAN classifiers for both ethnicities is depicted in Fig. \ref{ROC} and Table. \ref{CNN_GAN}. Looking at Fig. \ref{ROC}, for Africans, we reached an area under the curve (AUC) of 0.99 using the GAN classifier, while in Europeans, we achieved an AUC of 0.87 using the 3D CNN classifier. Table. \ref{CNN_GAN},  shows both classifiers can distinguish between FAS and control. The GAN classifier outperforms 3D CNN in all evaluation metrics. The GAN classifier in the African group reached the highest accuracy, sensitivity, and F1-score, while this classifier reached the highest specificity and precision in the European group.

\begin{table*}[ht]
\centering
\caption{Performance metrics for different lip segmentation methods. The best results for each metric and statistics are shown in bold. Interquartile range (IQR) }

\begin{tabular}{|c|c|c|c|c|c|c|c|}
\hline
Segmentation method & Statistic& Dice score (\%) & HD  & IoU (\%) & PA (\%) & PA$_C$ (\%)& VOE (\%)\\ \hline
\multirow{3}{*}{AUNet (RGB)} 
                   & Mean  & 71.54 & 3.69 & 56.37 & 99.53 & 81.04 &43.62 \\ \cline{2-8}
                   & Median  & 73.15 & 3.16 & 57.66 & 99.56 &82.98 &42.33  \\ \cline{2-8}
                   & IQR  & 8.74 & 1.17 & 10.80 & 0.17 &10.19 &10.80 \\ \hline
                   \hline
\multirow{3}{*}{AUNet ($\mathcal{T}$)}
                   & Mean & 80.09 & 3.56 & 67.80 & 99.69 &84.97 & 32.19  \\ \cline{2-8}
                   & Median & 83.23 & 2.24 & 71.27 & 99.74 & 88.22& 28.72 \\ \cline{2-8}
                   & IQR  & 6.98 & 1.00 & 10.09 & 0.12&6.81 & 10.10 \\ \hline
                   \hline
\multirow{3}{*}{S-AUNet (RGB)} 
                   & Mean  & 77.41 & 3.46 & 63.82 & 99.63 & 87.10& 36.17 \\ \cline{2-8}
                   & Median  & 79.21 & 3.00 & 65.58 & 99.66 &89.38 &34.41 \\ \cline{2-8}
                   & IQR  & 6.40 & 1.37 & \textbf{8.73} & 0.13 &7.55 &\textbf{8.73} \\ \hline
                   \hline 
\multirow{3}{*}{S-AUNet ($\mathcal{T}$)}
                   & Mean & \textbf{84.75} & \textbf{3.01} & \textbf{74.85} & \textbf{99.77}&\textbf{87.68} &\textbf{25.14} \\ \cline{2-8}
                   & Median  & \textbf{88.81} & \textbf{2.00} & \textbf{79.88} & \textbf{99.83} &\textbf{91.85} & \textbf{20.11}\\ \cline{2-8}
                   & IQR  & \textbf{5.71} & \textbf{0.82} & 9.07 & \textbf{0.11}& \textbf{5.91} & 9.08\\ \hline                   
\end{tabular}
\label{stat}
\end{table*}

\begin{table*}[ht]
\centering
\caption{Classification performance, A: African, EU: European. The best test accuracy is shown in bold.}

\begin{tabular}{|c|c|c|c|c|c|}
\hline
Classifier & Groups & Train loss & Train accuracy \tiny{$\%$} & Test loss & Test accuracy \tiny{$\%$} \\ \hline
\multirow{2}{*}{3D CNN} 
                   & A  & 0.0023 & 100 & 0.2061 & 95.65 \\ \cline{2-6}
                   & EU  & 0.0010 & 100 & 0.6016 & 90.56 \\ \hline
                   \hline
\multirow{2}{*}{GAN}
                   & A & 0.0357 & 99.27 & 0.1210 & \textbf{98.55} \\ \cline{2-6}
                   & EU  & 0.0795 & 97.73 & 0.5212 & 92.45 \\ \hline
\end{tabular}
\label{loss_acc}
\end{table*}

\begin{table*}[th!]
\centering
\caption{Classification metrics evaluated on the test dataset. The best results for each metric are shown in bold. }
\begin{tabular}{|c|c|c|c|c|c|}
\hline
\diagbox{Classifier - Group}{Metric} & Accuracy \tiny{$\%$} & Sensitivity \tiny{$\%$} & Specificity \tiny{$\%$} & Precision \tiny{$\%$} & F1-score \tiny{$\%$} \\ \hline
3D CNN - A &  95.65     &  87.50     & 96.72      &  77.78     & 82.35      \\ \hline
3D CNN - EU &  90.56     & 61.54      & 95.58      &   70.59    &   65.75    \\ \hline
GAN - A & \textbf{98.55}      & \textbf{100}      & 98.36      & 88.89      &  \textbf{94.12}   \\ \hline
GAN - EU & 92.45      & 53.85      &  \textbf{99.12}     &   \textbf{91.30}    &   67.74    \\ \hline
\end{tabular}
\label{CNN_GAN}
\end{table*}

\section{Discussion} \label{discussion}
This paper introduces a methodology to segment upper lips in facial images with a clinical application for FAS facial assessment. In the first part, we presented a multidimensional attention UNet-based method for upper lip segmentation. As illustrated in Fig. \ref{4segmentation_w} and Fig. \ref{4segmentation_AA}, using LBP and GLBP facilitates revealing micro-patterns on the upper lip, both on the upper and lower parts. This approach improves the robustness of the lip segmentation method against image quality, skin tones, brightness, and image contrast. In addition, integrating the multidimensional input and the sequential segmentation process facilitates refining the boundaries of the upper lips. We support this conclusion by thoroughly analyzing the assessment metrics depicted in Table. \ref{stat}. More specifically, comparing the pixel accuracy per class and IoU between AUNet (RGB) and AUNet ($\mathcal{T}$) reveals that multidimensional input enhances lip contour reconstruction with higher accuracy than utilizing only raw RGB images. Furthermore, comparing S-AUNet (RGB) and S-AUNet ($\mathcal{T}$), the sequential approach helps refine boundaries, especially around Cupid’s bow.

In the second part of the study, utilizing segmented upper lips, we implemented two classifiers, 3D CNN and GAN, to identify those with FAS using latent space. FAS identification using lip analysis has been explored in prior studies. Suttie et al. \cite{suttie2013facial} investigated several contributing characteristics in FAS identification including perioral (mouth region), perinasal (nasal region), periorbital (eye orbit region), profile, and the full face features in a mixed ancestry population. They reached an accuracy of $88.30\%$ in distinguishing between FAS/pFAS and healthy controls using an integration of perioral features with any of the closest mean (CM), linear discriminant analysis (LDA), or support vector machines (SVM). Notably, perioral features encompass all the characteristics of the mouth and the surrounding area, with a thin upper lip being one of these features. Another similar study has been done by Suttie et al. \cite{suttie2017facial} to build ethnicity-specific models for FAS identification. They reached an accuracy of $84.0\%$ and $73.0\%$ for African and European populations using lip vermilion and an LDA classifier on principal components representing 3D shape. Compared  with the previous studies, we improved FAS identification to $98.55\%$ in Africans and $92.45\%$ in Europeans, utilizing latent space and the GAN classifier.

\begin{figure*}[t]
    \centering
    \includegraphics[width=1\textwidth]{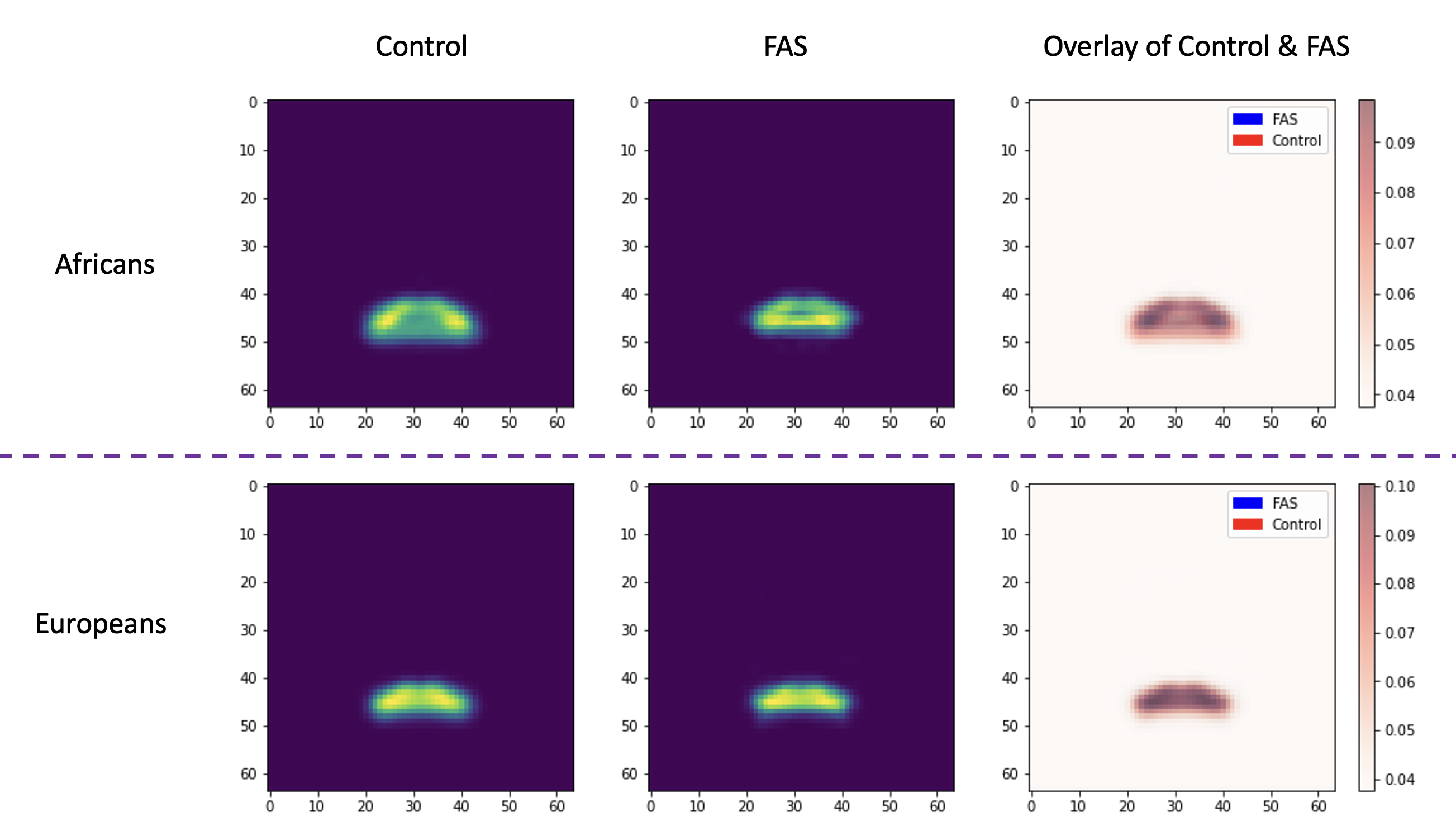}
    \caption{Comparing 2D latent representation between control and FAS.}
    \label{latent}
\end{figure*}

Beyond the classification method, latent space was successful when independently analyzed to identify FAS. Using an autoencoder, we can compress the segmented images into a common feature space while maintaining key features. To further investigate the analysis, we projected 3D latent into 2D latent by averaging them across the z-axis. In the next step, for each group (i.e., control or FAS), we computed the average 2D latent by averaging across all subjects per group. As illustrated in Fig. \ref{latent}, for both ethnicities (that is, Africans and Europeans), the thickness and area of the lips decreased in FAS compared to the control. The decrease in lip thickness is more pronounced in Africans than in Europeans. Therefore, calculating the distance between the 2D latent of an individual and the average 2D latent of the control or FAS groups could be regarded as a beneficial tool for assessing FAS in the clinical environment.

\subsection{Limitations and Future Work}
This paper proposes a segmentation method that utilizes segmentation masks in the training phase. To generate the segmentation masks, we used annotated anatomical landmarks as initial points and estimated the intermediate points by the method explained in Section. \ref{mask}. However, this approach made the method semi-supervised, meaning it relies on the initial anatomical landmarks to generate the lip masks in the training phase. An alternative way to find the initial anatomical landmarks involves using self-supervised methods to estimate the key points. Therefore, to develop an unsupervised lip segmentation method, we propose integrating self-supervised key point detection methods, such as \cite{detone2018superpoint}, with the mask generation method proposed in our paper. 

Furthermore, using a supervised method, we used segmented lips to assess FAS, where the FAS status is known. As mentioned in Section \ref{FAS}, three facial cardinal features are involved in FAS diagnoses, with a thin upper lip being one of them. Since the current scoring system for lip thickness is based on a qualitative comparison with the 5-point Likert scale, it introduces subjectivity that could cause inconsistency in the control or FAS labels. To overcome this limitation, as part of future work, we propose to develop an unsupervised method first to evaluate the thickness of the lip and then use it to assess FAS. This approach could lead to a more objective method less reliant on the clinician's visual assessment.
\section{Conclusion}\label{conclusion}
This paper proposes an upper lip segmentation method using a multidimensional attention UNet-based approach. Integrating multidimensional inputs, sequential UNets, and estimated masks facilitates extracting micro-patterns, refining the boundaries, focusing on regional information, and improving the robustness of the lip segmentation method to image quality, skin tones, and lighting. The proposed method reached a mean dice score of $84.75\%$, and a mean of pixel accuracy of $99.77\%$ in upper lip segmentation. A further analysis was conducted by integrating the lip segmentation method and the GAN classifier to identify those with FAS, resulting in $98.55\%$ accuracy in the African group. In the future, we will focus on building an unsupervised method for both mask generation and FAS identification, thereby making the whole process user-independent.

\section*{Data availability}
Restrictions apply to the availability of these data, which were used under license for the current study from the CIFASD consortium and are not publicly available. Identifiable human data are available to CIFASD members after securing ethical permission and appropriate data access agreements.
\section*{Ethics statement}
Data were obtained from the Collaborative Initiative on Fetal Alcohol Spectrum Disorders (CIFASD) consortium (https://cifasd.org/data-sharing) from multiple sites across the USA (San Diego, Minnesota, and Atlanta) and South Africa, which have individual ethical approvals in place. Parents or guardians of participants provided consent for image collection and FASD-dysmorphology assessments at each site. Data-sharing agreements were established as part of the CIFASD consortium, and ethical approval for this study was reviewed and approved in June 2022  by the Oxford Tropical Research Ethics Committee (OxTREC), University of Oxford. OxTREC reference for this approval is 519-17.
\section*{Declaration of competing interest}
None declared.

\section*{ACKNOWLEDGMENT}
All of this work was done in conjunction with the Collaborative Initiative on Fetal Alcohol Spectrum Disorders (CIFASD), which is funded by grants from the National Institute on Alcohol Abuse and Alcoholism (NIAAA). Additional information about CIFASD can be found at \cite{cifasd}. This work was supported by NIH grants U01AA014809 (M.S). Data were obtained from the CIFASD consortium
(https://cifasd.org/data-sharing) from multiple sites across the
USA (San Diego, Minnesota, and Atlanta) and from the
Surrey and Borders Partnership NHS Foundation Trust UK
FASD Clinic (United Kingdom) (R.M). CIFASD data collection
sites were supported by NIAAA grants: U01AA014835
(C.C), U01AA014834 (S.N.M), U01AA026102 (J.R.W),
U01AA030164 (J.R.W), and U01AA026108 (C.D.C).


\bibliographystyle{IEEEtran}

\begin{thebibliography}{10}

\bibitem{rot2018deep}
Peter Rot, {\v{Z}}iga Emer{\v{s}}i{\v{c}}, Vitomir Struc, and Peter Peer.
\newblock Deep multi-class eye segmentation for ocular biometrics.
\newblock In {\em 2018 IEEE international work conference on bioinspired intelligence (IWOBI)}, pages 1--8. IEEE, 2018.

\bibitem{dibekliouglu2009nasal}
Hamdi Dibeklio{\u{g}}lu, Berk G{\"o}kberk, and Lale Akarun.
\newblock Nasal region-based 3d face recognition under pose and expression variations.
\newblock In {\em Advances in Biometrics: Third International Conference, ICB 2009, Alghero, Italy, June 2-5, 2009. Proceedings 3}, pages 309--318. Springer, 2009.

\bibitem{liew2003segmentation}
AW-C Liew, Shu~Hung Leung, and Wing~Hong Lau.
\newblock Segmentation of color lip images by spatial fuzzy clustering.
\newblock {\em IEEE transactions on Fuzzy Systems}, 11(4):542--549, 2003.

\bibitem{ryu2005improving}
Jong-Seong Ryu, Sun-Gyoo Park, Taek-Jong Kwak, Min-Youl Chang, Moon-Eok Park, Khee-Hwan Choi, Kyung-Hye Sung, Hyun-Jong Shin, Cheon-Koo Lee, Yun-Seok Kang, et~al.
\newblock Improving lip wrinkles: lipstick-related image analysis.
\newblock {\em Skin Research and Technology}, 11(3):157--164, 2005.

\bibitem{ma2024decoupled}
Boyao Ma, Yuanping Cao, and Lei Zhang.
\newblock Decoupled two-stage talking head generation via gaussian-landmark-based neural radiance fields.
\newblock {\em Authorea Preprints}, 2024.

\bibitem{sheng2024deep}
Changchong Sheng, Gangyao Kuang, Liang Bai, Chenping Hou, Yulan Guo, Xin Xu, Matti Pietik{\"a}inen, and Li~Liu.
\newblock Deep learning for visual speech analysis: A survey.
\newblock {\em IEEE Transactions on Pattern Analysis and Machine Intelligence}, 2024.

\bibitem{suttie2013facial}
Michael Suttie, Tatiana Foroud, Leah Wetherill, Joseph~L Jacobson, Christopher~D Molteno, Ernesta~M Meintjes, H~Eugene Hoyme, Nathaniel Khaole, Luther~K Robinson, Edward~P Riley, et~al.
\newblock Facial dysmorphism across the fetal alcohol spectrum.
\newblock {\em Pediatrics}, 131(3):e779--e788, 2013.

\bibitem{dixon2011cleft}
Michael~J Dixon, Mary~L Marazita, Terri~H Beaty, and Jeffrey~C Murray.
\newblock Cleft lip and palate: understanding genetic and environmental influences.
\newblock {\em Nature Reviews Genetics}, 12(3):167--178, 2011.

\bibitem{eveno2001new}
Nicolas Eveno, Alice Caplier, and P-Y Coulon.
\newblock New color transformation for lips segmentation.
\newblock In {\em 2001 IEEE Fourth Workshop on Multimedia Signal Processing (Cat. No. 01TH8564)}, pages 3--8. IEEE, 2001.

\bibitem{guan2008automatic}
Y-P Guan.
\newblock Automatic extraction of lips based on multi-scale wavelet edge detection.
\newblock {\em IET Computer Vision}, 2(1):23--33, 2008.

\bibitem{lip}
Lip, 2025.
\newblock Retrieved on February 24, 2025.

\bibitem{liew2000lip}
Alan Wee-Chung Liew, Shu~Hung Leung, and Wing~Hong Lau.
\newblock Lip contour extraction using a deformable model.
\newblock In {\em Proceedings 2000 International Conference on Image Processing (Cat. No. 00CH37101)}, volume~2, pages 255--258. IEEE, 2000.

\bibitem{shdaifat2003active}
Islam Shdaifat, R~Grigat, and Detlev Langmann.
\newblock Active shape lip modeling.
\newblock In {\em Proceedings 2003 International Conference on Image Processing (Cat. No. 03CH37429)}, volume~2, pages II--875. IEEE, 2003.

\bibitem{delmas1999automatic}
Patrice Delmas, Pierre-Yves Coulon, and Vincent Fristot.
\newblock Automatic snakes for robust lip boundaries extraction.
\newblock In {\em 1999 IEEE International Conference on Acoustics, Speech, and Signal Processing. Proceedings. ICASSP99 (Cat. No. 99CH36258)}, volume~6, pages 3069--3072. IEEE, 1999.

\bibitem{long2015fully}
Jonathan Long, Evan Shelhamer, and Trevor Darrell.
\newblock Fully convolutional networks for semantic segmentation.
\newblock In {\em Proceedings of the IEEE conference on computer vision and pattern recognition}, pages 3431--3440, 2015.

\bibitem{ronneberger2015u}
Olaf Ronneberger, Philipp Fischer, and Thomas Brox.
\newblock U-net: Convolutional networks for biomedical image segmentation.
\newblock In {\em Medical image computing and computer-assisted intervention--MICCAI 2015: 18th international conference, Munich, Germany, October 5-9, 2015, proceedings, part III 18}, pages 234--241. Springer, 2015.

\bibitem{oktay2018attention}
Ozan Oktay, Jo~Schlemper, Loic~Le Folgoc, Matthew Lee, Mattias Heinrich, Kazunari Misawa, Kensaku Mori, Steven McDonagh, Nils~Y Hammerla, Bernhard Kainz, et~al.
\newblock Attention u-net: Learning where to look for the pancreas.
\newblock {\em arXiv preprint arXiv:1804.03999}, 2018.

\bibitem{hoyme2016updated}
H~Eugene Hoyme, Wendy~O Kalberg, Amy~J Elliott, Jason Blankenship, David Buckley, Anna-Susan Marais, Melanie~A Manning, Luther~K Robinson, Margaret~P Adam, Omar Abdul-Rahman, et~al.
\newblock Updated clinical guidelines for diagnosing fetal alcohol spectrum disorders.
\newblock {\em Pediatrics}, 138(2), 2016.

\bibitem{astley2015palpebral}
Susan~J Astley.
\newblock Palpebral fissure length measurement: accuracy of the fas facial photographic analysis software and inaccuracy of the ruler.
\newblock {\em J Popul Ther Clin Pharmacol}, 22(1):e9--e26, 2015.

\bibitem{ojala1996comparative}
Timo Ojala, Matti Pietik{\"a}inen, and David Harwood.
\newblock A comparative study of texture measures with classification based on featured distributions.
\newblock {\em Pattern recognition}, 29(1):51--59, 1996.

\bibitem{ahonen2006face}
Timo Ahonen, Abdenour Hadid, and Matti Pietikainen.
\newblock Face description with local binary patterns: Application to face recognition.
\newblock {\em IEEE transactions on pattern analysis and machine intelligence}, 28(12):2037--2041, 2006.

\bibitem{ojala2002multiresolution}
Timo Ojala, Matti Pietikainen, and Topi Maenpaa.
\newblock Multiresolution gray-scale and rotation invariant texture classification with local binary patterns.
\newblock {\em IEEE Transactions on pattern analysis and machine intelligence}, 24(7):971--987, 2002.

\bibitem{chao2015facial}
Wei-Lun Chao, Jian-Jiun Ding, and Jun-Zuo Liu.
\newblock Facial expression recognition based on improved local binary pattern and class-regularized locality preserving projection.
\newblock {\em Signal Processing}, 117:1--10, 2015.

\bibitem{moghaddasi2016automatic}
Hanie Moghaddasi and Saeed Nourian.
\newblock Automatic assessment of mitral regurgitation severity based on extensive textural features on 2d echocardiography videos.
\newblock {\em Computers in biology and medicine}, 73:47--55, 2016.

\bibitem{moghaddasi2022classification}
Hanie Moghaddasi, Richard~C Hendriks, Alle-Jan van~der Veen, Natasja~MS de~Groot, and Borb{\'a}la Hunyadi.
\newblock Classification of de novo post-operative and persistent atrial fibrillation using multi-channel ecg recordings.
\newblock {\em Computers in Biology and Medicine}, 143:105270, 2022.

\bibitem{moghaddasi2024model}
Hanie Moghaddasi.
\newblock Model-based feature engineering of atrial fibrillation.
\newblock 2024.

\bibitem{yazid2020variable}
Muhammad Yazid and Mahrus~Abdur Rahman.
\newblock Variable step dynamic threshold local binary pattern for classification of atrial fibrillation.
\newblock {\em Artificial Intelligence in Medicine}, 108:101932, 2020.

\bibitem{unay2007robustness}
Devrim Unay, Ahmet Ekin, Mujdat Cetin, Radu Jasinschi, and Aytul Ercil.
\newblock Robustness of local binary patterns in brain mr image analysis.
\newblock In {\em 2007 29th Annual International Conference of the IEEE Engineering in Medicine and Biology Society}, pages 2098--2101. IEEE, 2007.

\bibitem{fu2021facial}
Zeyu Fu, Jianbo Jiao, Michael Suttie, and J~Alison Noble.
\newblock Facial anatomical landmark detection using regularized transfer learning with application to fetal alcohol syndrome recognition.
\newblock {\em IEEE journal of biomedical and health informatics}, 26(4):1591--1601, 2021.

\bibitem{kim2020automatic}
Yoon-Chul Kim, Khu~Rai Kim, and Yeon~Hyeon Choe.
\newblock Automatic myocardial segmentation in dynamic contrast enhanced perfusion mri using monte carlo dropout in an encoder-decoder convolutional neural network.
\newblock {\em Computer methods and programs in biomedicine}, 185:105150, 2020.

\bibitem{shaziya2018automatic}
Humera Shaziya, K~Shyamala, and Raniah Zaheer.
\newblock Automatic lung segmentation on thoracic ct scans using u-net convolutional network.
\newblock In {\em 2018 International conference on communication and signal processing (ICCSP)}, pages 0643--0647. IEEE, 2018.

\bibitem{salehi2017auto}
Seyed Sadegh~Mohseni Salehi, Deniz Erdogmus, and Ali Gholipour.
\newblock Auto-context convolutional neural network (auto-net) for brain extraction in magnetic resonance imaging.
\newblock {\em IEEE transactions on medical imaging}, 36(11):2319--2330, 2017.

\bibitem{hinton2006reducing}
Geoffrey~E Hinton and Ruslan~R Salakhutdinov.
\newblock Reducing the dimensionality of data with neural networks.
\newblock {\em science}, 313(5786):504--507, 2006.

\bibitem{ji20123d}
Shuiwang Ji, Wei Xu, Ming Yang, and Kai Yu.
\newblock 3d convolutional neural networks for human action recognition.
\newblock {\em IEEE transactions on pattern analysis and machine intelligence}, 35(1):221--231, 2012.

\bibitem{goodfellow2014generative}
Ian Goodfellow, Jean Pouget-Abadie, Mehdi Mirza, Bing Xu, David Warde-Farley, Sherjil Ozair, Aaron Courville, and Yoshua Bengio.
\newblock Generative adversarial nets.
\newblock {\em Advances in neural information processing systems}, 27, 2014.

\bibitem{suttie2017facial}
Michael Suttie, Leah Wetherill, Sandra~W Jacobson, Joseph~L Jacobson, H~Eugene Hoyme, Elizabeth~R Sowell, Claire Coles, Jeffrey~R Wozniak, Edward~P Riley, Kenneth~L Jones, et~al.
\newblock Facial curvature detects and explicates ethnic differences in effects of prenatal alcohol exposure.
\newblock {\em Alcoholism: Clinical and Experimental Research}, 41(8):1471--1483, 2017.

\bibitem{detone2018superpoint}
Daniel DeTone, Tomasz Malisiewicz, and Andrew Rabinovich.
\newblock Superpoint: Self-supervised interest point detection and description.
\newblock In {\em Proceedings of the IEEE conference on computer vision and pattern recognition workshops}, pages 224--236, 2018.

\bibitem{cifasd}
CIFASD.
\newblock Collaborative initiative on fetal alcohol spectrum disorders, 2025.
\newblock Retrieved on February 24, 2025.

\end{thebibliography}

\end{document}